\documentclass[twoside,a4paper,11pt]{article}
\usepackage{times}
\usepackage{elcvia}
\usepackage[english]{babel}
\usepackage{graphicx}
\usepackage{amsmath}
\usepackage{url}
\usepackage{booktabs}
\usepackage{float}
\usepackage{titlesec}
\usepackage{enumitem}
\usepackage{parskip}
\setlist[itemize]{noitemsep, topsep=0pt}
\titlespacing*{\subsection}{0pt}{1ex plus 0.5ex minus .2ex}{0.5ex plus .2ex}
\titlespacing*{\section}{0pt}{2ex plus 0.5ex minus .2ex}{1ex plus .2ex}

\begin{document}
\begin{center}
{\fontsize{20pt}{24pt}\selectfont \textbf{LipLingo: CNN Model for Lip Reading}}\\[1em]
Mahmoud S. Mayaleh$^{1}$,
Liban Ali Achkir$^{2}$,
Nijat Mehdiyev$^{2}$,
Ahmet Özdil$^{2}$ \\[0.5em]
$^{1}$National Conservatory of Arts and Crafts (CNAM), Paris, France \\ 
$^{2}$Ostim Technical University, Ankara, Türkiye \\[1em]
\texttt{mahmoud.mayaleh@lecnam.net, liban.ali.achkir@ostimteknik.edu.tr,}\\
\texttt{240910604@ostimteknik.edu.tr, ahmet.ozdil@ostimteknik.edu.tr}
\end{center}


\begin{abstract}
Lip reading plays a crucial role in applications such as speech recognition, assistive technologies for the hearing-impaired, and biometric authentication. However, performance often degrades when the speaker varies and when certain speech sounds are visually similar. To address these challenges, we propose LipLingo, a robust lip-reading system based on the LipNet architecture, enhanced with a standardized preprocessing pipeline for consistent mouth-region representation, a rotational validation strategy for improved generalization to unseen speakers, and a flexible similarity evaluation for more reliable sentence-level assessment. The model architecture combines 3D convolutional layers for spatio-temporal feature extraction and bidirectional recurrent layers for sequence modeling, optimized using the Adam optimizer and Connectionist Temporal Classification (CTC) loss. On the GRID dataset, LipLingo achieved 95.0\% character-level accuracy (CER 4.99\%) and 87.2\% sentence-level accuracy. At the corpus level, the system obtained a (WER 13.1\%), outperforming the baseline LipNet and demonstrating competitive performance relative to recent 3D-ResNet + Bi-GRU + MixUp approaches. These results provide insights into strategies that enhance lip-reading robustness and highlight directions for future work toward real-world deployment.
\end{abstract}

\section{Introduction}

\subsection{General Overview of Lip Reading}
Human language understanding relies on a combination of auditory and visual cues. 
While spoken language is primarily transmitted via sound waves, visual information---particularly lip movements---provides complementary data that improves comprehension. 
Lip reading, also referred to as \emph{speech reading}, is the process of interpreting spoken words by observing these visual cues. 
It has several practical applications, including:
\begin{itemize}
    \item \textbf{Speech Recognition:} Enhances transcription accuracy when integrated with audio-based systems.
    \item \textbf{Assistive Technologies:} Enables hearing-impaired individuals to better understand spoken communication.
    \item \textbf{Biometric Authentication:} Utilizes unique lip movement patterns for speaker verification.
\end{itemize}

Despite recent advances in machine learning, lip reading systems face challenges due to variations in facial features, illumination conditions, and the visual similarity of certain phonemes. 
These factors complicate accurate interpretation and limit performance in real-world scenarios.

\subsection{A Deep Dive into Improved Speech Reading}
This study explores deep learning-based approaches to automatic lip reading, aiming to achieve higher accuracy and improved generalization compared with existing pre-trained systems. 
Our approach is based on the LipNet architecture~\cite{assael2016lipnet}, which has shown strong performance on sentence-level lip reading tasks.

Recent advances in lip reading are based on modern sequence models and pre-training. Transformer/Conformer encoders now have very low error rates on large, in-the-wild benchmarks such as LRS3 \cite{chang2023conformers}, and self-supervised methods (e.g., AV-HuBERT) learn strong visual speech features with far less labeled data \cite{shi2022avhubert}. Landmark-guided transformers and efficient windowed-attention encoders further improve generalization to new speakers and reduce computational effort \cite{xue2023lipformer, park2025swinlip}. Unified audio–visual models can also run in audio-only\cite{haliassos2024unified}, video-only or fused modes, improving robustness when one modality is unreliable.

Most of the above systems rely on very large, diverse corpora and significant computational power; they are not directly comparable on GRID’s controlled grammar and scale \cite{cooke2006grid}. Therefore, we use LipNet as a reproducible, end-to-end baseline on GRID so that we can isolate the effects of our contributions without confounding factors from large-scale pre-training, while discussing broader trends in the literature.

\subsection{Our Specific Objectives}
This work is guided by four primary objectives:
\begin{enumerate}
    \item Evaluate the potential of the LipNet architecture for customized training on our dataset.
    \item Develop a tailored model that integrates additional preprocessing and architectural enhancements.
    \item Conduct rigorous performance evaluation and compare against baseline and state-of-the-art models.
    \item Identify key factors that influence accuracy and robustness to guide future improvements.
\end{enumerate}

While LipNet and similar architectures achieve good performance under constrained settings, their generalization to diverse speakers and environments remains limited. This study focuses on addressing these challenges by proposing improvements that enhance model robustness and practical usability, which is crucial for assistive technologies and biometric applications.

\section{Literature Review}

\subsection{Overview of Related Work in Lip Reading}
Lip reading, also referred to as \emph{speech reading}, involves decoding spoken language by interpreting the visible movements of a speaker's lips and face. 
The advent of deep learning has significantly improved the ability of models to learn complex spatio-temporal representations. 
Earlier research focused primarily on recognizing isolated words or limited vocabularies. 
Recent efforts, however, have shifted toward sentence-level transcription, speaker-independent recognition, and end-to-end architectures that minimize the need for manual feature engineering.

\subsection{Existing Models, Techniques, and Datasets}

\textbf{LipNet.} 
Assael et al.~\cite{assael2016lipnet} introduced LipNet, a pioneering end-to-end model for sentence-level lip reading that set a new benchmark on the GRID dataset. 
LipNet combines spatio-temporal convolutional layers to capture both spatial features (e.g., lip shape) and temporal dynamics (e.g., mouth movements across frames). 
These features are then processed by Long Short-Term Memory (LSTM) layers~\cite{hochreiter1997lstm}, which model the sequential nature of speech. 
The use of Connectionist Temporal Classification (CTC) loss~\cite{graves2006ctc} enables alignment-free mapping between video frames and character sequences. 
LipNet achieved significant performance gains over previous approaches and inspired numerous follow-up studies.

\textbf{3D Convolutional Neural Networks.} 
Exarchos et al.~\cite{exarchos2024lipreading} proposed a 3D CNN architecture capable of modeling both spatial and temporal dimensions simultaneously. 
Unlike 2D CNNs, which process individual frames independently, 3D convolutions operate across sequences of frames, enabling the network to capture motion dynamics inherent to speech. 
Some variants further incorporate recurrent layers, such as LSTMs or Gated Recurrent Units (GRUs)~\cite{cho2014gru}, to enhance temporal modeling. 
While powerful, 3D CNNs are computationally demanding and prone to overfitting, particularly on smaller datasets.

\textbf{Viseme-Based Recognition with Hidden Markov Models.} 
Earlier approaches often relied on hand-crafted features and statistical models.
Lu et al.~\cite{lu2006visemes} and Baum and Petrie~\cite{baum1966hmm} developed systems for classifying \emph{visemes}, which are visually distinct clusters of phonemes~\cite{petridis2018audiovisual}. 
Hidden Markov Models (HMMs) modeled the probabilistic transitions between visemes over time. 
Although viseme-based recognition simplified classification, it lacked the fine-grained detail required for accurate word recognition in continuous speech.

\textbf{Deep Autoencoders.} 
Zhou and Paffenroth~\cite{zhou2017autoencoders} investigated stacked denoising autoencoders to learn hierarchical feature representations without requiring labeled data. 
These models compress and reconstruct input video sequences, learning robust encodings even in noisy conditions. 
Unsupervised pre-training can be particularly valuable for low-resource datasets, although autoencoders typically require additional sequence-labeling components for complete speech recognition pipelines.

While LipNet set an important benchmark, it also has clear limitations. Its training relies on the GRID dataset’s constrained grammar and vocabulary, which makes results less representative of natural speech. In addition, LipNet struggles with visually similar phonemes (e.g., 'p' vs. 'b') and has limited robustness to speaker variation or environmental changes. Later architectures such as 3D CNNs and ResNet–BiGRU hybrids improve spatio-temporal feature capture, but at the cost of higher computational complexity and larger datasets. Transformer-based methods (e.g., Conformer VSR, LipFormer) show further accuracy gains, yet depend on large-scale pretraining and benchmark datasets (LRS2 \cite{chung2017lrs2}, LRS3), which are not always practical. These large-scale datasets contain tens to hundreds of hours of unconstrained speech from diverse speakers, but they demand substantial computational resources (multi-GPU training for weeks) that are not accessible in many research or applied settings, they introduce licensing and storage constraints, since distributing and processing hundreds of gigabytes is not feasible for all institutions. For these reasons, when focusing on controlled datasets such as GRID, using transformer-based systems pre-trained on LRS2/LRS3 may be excessive or misaligned with the study goals. LipLingo addresses part of the gap by proposing a standardized preprocessing pipeline that reduces speaker-specific variability and a rotational validation strategy to improve generalization on GRID, while keeping the model lightweight.

\textbf{Datasets.} 
The GRID audiovisual sentence corpus remains one of the most widely used benchmarks for lip reading research. 
Its structured sentence format and consistent recording conditions support reproducible experiments, yet its limited speaker diversity and restricted vocabulary hinder generalization to natural, unconstrained speech.

Beyond GRID, the field now evaluates models on more diverse datasets such as LRW (word-level, 500 classes, short clips) \cite{chung2016lrw}, LRW-1000 (large-scale, naturally distributed, Mandarin) \cite{yang2019lrw1000}, and LRS2/LRS3 (sentence-level, unconstrained speech from BBC/TED). These benchmark datasets introduce challenges such as large vocabulary, varying sentence structures, and uncontrolled recording conditions. They highlight the gap between controlled benchmarks like GRID and real-world deployment. In this study, we focus on GRID for reproducibility and fair comparison with LipNet, but we acknowledge the importance of testing on larger benchmark datasets in future work.

\begin{table}[h!]
\centering
\caption{Sentence-level lip-reading models and their reported performance (WER-focused).}
\begin{tabular}{|l|l|l|}
\hline
\textbf{Model} & \textbf{Dataset} & \textbf{Performance} \\ \hline
LipNet (2016) & GRID & WER 11.4\% \\
3D ResNet + Bi-GRU + MixUp (2024) & GRID & WER 11.3\%\\
Conformer VSR (2023) & LRS3 (TED) & WER 12.8\%\\
SwinLip (2025) & LRS2 & WER 37.01\% \\
AV-HuBERT (2022) & LRS2 / LRS3 & WER 26.9\%  \\
\textbf{LipLingo (ours)} & \textbf{GRID} & \textbf{WER 13.1\%} \\ \hline
\end{tabular}
\end{table}

\subsection{Gaps in Current Research}
Despite substantial progress, several challenges persist:
\begin{itemize}
    \item \textbf{Vocabulary Scale and Speaker Independence:} Many models achieve high accuracy on constrained datasets but degrade when applied to unconstrained speech or unseen speakers. Achieving true speaker independence remains an open problem~\cite{oghbaie2021advances}.
    \item \textbf{Environmental Robustness:} Variations in lighting, occlusions (e.g., facial hair, masks), and speaking styles significantly affect performance. Data augmentation and multimodal sensor fusion have been proposed as mitigations but require further validation.
    \item \textbf{Computational accessibility:} Transformer-based and self-supervised models achieve state-of-the-art results, but they require large-scale datasets (LRS2, LRS3, LRW-1000) and weeks of multi-GPU training, making them impractical for smaller labs or real-world deployments where compute and data are limited.
    \item \textbf{Interpretability and Transparency:} Deep learning architectures often function as black boxes, providing limited insight into decision-making. Incorporating domain knowledge, such as biomechanical models of speech production, may improve interpretability.
\end{itemize}

Given these limitations, our work builds upon LipNet but introduces enhanced preprocessing steps and architectural refinements designed to improve accuracy, robustness, and real-world applicability.

\section{Methodology}

The complete implementation, including preprocessing pipelines, model training scripts, and evaluation tools, is available at our public repository: \url{https://github.com/mahmoudmayaleh/LipLingo}.

\subsection{Dataset}
The experiments were conducted using the GRID audiovisual sentence corpus~\cite{cooke2006grid}, which consists of video and audio recordings from 34 speakers (18 male, 16 female), each uttering 1,000 sentences, resulting in a total of 34,000 samples. 
The sentences follow a fixed syntactic structure (e.g., ``place red at G9 now''), which facilitates supervised learning while providing speaker and articulation variability.

\subsection{Data Preprocessing}
\begin{figure}[H]
 \centering
 \includegraphics[width=1\linewidth]{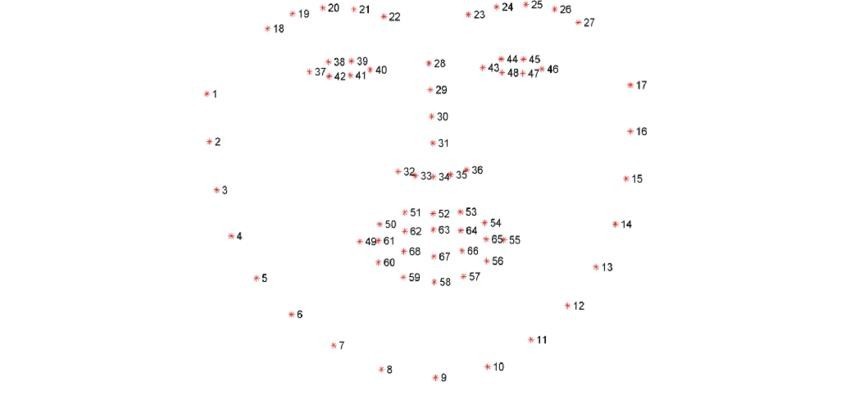}
 \caption{Facial landmark detection using MediaPipe for isolating the mouth region.~\cite{lugaresi2019mediapipe}.}
 \label{fig:mediapipe_landmarks}
\end{figure}

Each video was decomposed into individual frames using OpenCV~\cite{opencvcolorspace}. 
Face detection and extraction of 468 facial landmarks were performed using MediaPipe~\cite{lugaresi2019mediapipe}. 
A bounding box around the mouth region was computed based on these landmarks to isolate relevant visual features and minimize background noise (Figure~\ref{fig:mediapipe_landmarks}).  

\begin{figure}[H]
    \centering
    \includegraphics[width=0.8\linewidth]{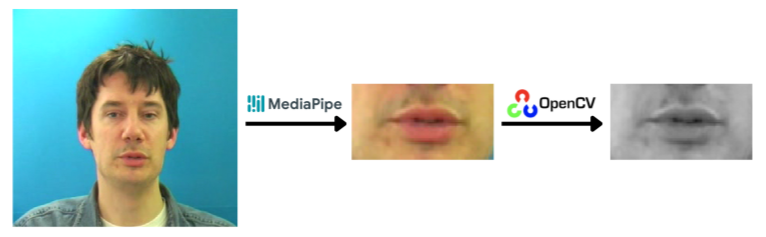}
    \caption{Example of mouth region extraction and grayscale conversion used in preprocessing. This step standardizes input features and reduces color-related variability, improving model robustness to lighting differences.}
    \label{fig:mouth_crop}
\end{figure}

The extracted mouth regions were resized to $100 \times 50$ pixels to standardize input dimensions. 
To reduce computational complexity, frames were converted to grayscale:
\begin{equation}
Y = 0.299R + 0.587G + 0.114B.
\end{equation}
Sequences shorter than the maximum length were zero-padded, and pixel intensities were normalized by subtracting the mean and dividing by the standard deviation. Examples of preprocessed frames are shown in Figure~\ref{fig:mouth_crop}.

Although the GRID dataset was collected under controlled laboratory conditions with minimal background noise and stable illumination, we did not perform any explicit denoising.
Our preprocessing pipeline (normalization, padding, grayscale conversion) standardizes the inputs across all sessions, but does not simulate environmental variability such as occlusions, motion blur or background noise.
These factors are considered limitations of the present study and will be addressed in future work through targeted data augmentation and multimodal fusion strategies.

\subsection{Data Pipeline}
The TensorFlow \texttt{tf.data} API~\cite{abadi2015tensorflow} was employed to construct an efficient data pipeline:  
\begin{itemize}
 \item Loading of video frame sequences and alignment labels.
 \item Application of preprocessing functions in parallel.
 \item Batching of samples with automatic sequence padding.
 \item Prefetching to optimize GPU utilization.
\end{itemize}
This design ensured consistently formatted and efficiently streamed data during training.

\subsection{Model Architecture}
\begin{figure}[H]
    \centering
    \includegraphics[width=0.85\linewidth]{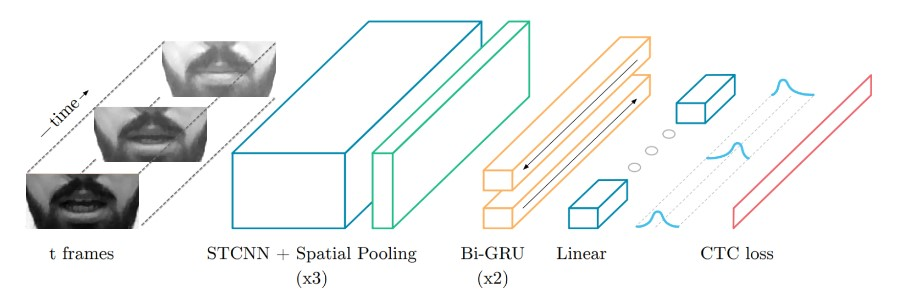}
    \caption{Lip reading model architecture combining CNN, GRU, and CTC loss.}
    \label{fig:model_architecture}
\end{figure}

The LipLingo model architecture extends LipNet~\cite{assael2016lipnet} with modifications to improve feature extraction and sequence modeling. 
As shown in (Figure~\ref{fig:model_architecture}), the architecture includes three 3D convolutional layers, two bidirectional GRU layers, and a final dense-softmax output layer optimized with CTC loss.  
\begin{itemize}
 
 \item \textbf{Grayscale input.} 
All video frames were converted to grayscale before being forwarded to the network.
This choice reduces computational costs and memory usage while standardizing the input for all speakers and recording conditions.
Although this step removes potentially informative color cues (e.g., tongue visibility, contrast between teeth and lips), previous work has shown that lip contour and motion provide the dominant features for visual speech recognition on datasets such as GRID. We acknowledge this trade-off and note that the inclusion of color information is a possibility for future work.

 \item \textbf{3D Convolutional layers.}
Three spatio-temporal convolutional layers were used to capture both spatial lip features and temporal dynamics across consecutive frames.
This depth was selected as a balance: Deeper stacks provided only marginal improvements in accuracy but resulted in overfitting and increased computational costs, while shallower models underfit the task.

 \item \textbf{Bidirectional GRU Layers:} Two bidirectional GRU layers model sequential dependencies in forward and backward directions~\cite{cho2014gru}. GRUs were chosen instead of LSTMs to reduce the number of parameters and training costs while maintaining a strong temporal modeling capacity. Experience has shown that adding more recurrent layers does not improve the accuracy on GRID and increases training instability, so the architecture was fixed at two GRU layers.

\item \textbf{Dense and SoftMax Layers:} A fully connected (dense) layer projects the recurrent features onto the character class space.
 This is followed by a SoftMax activation, that generates a probability distribution over all possible characters at each time step.
 This design allows the model to output flexible sequence predictions aligned with the CTC loss, while keeping the final layer lightweight to avoid overfitting on the relatively small GRID dataset.
\end{itemize}

\subsection{Connectionist Temporal Classification (CTC) Loss}
Because the alignment between input frames and output characters is unknown, the Connectionist Temporal Classification (CTC) loss~\cite{graves2006ctc} was adopted:
\begin{equation}
\mathcal{L}_{\mathrm{CTC}} = -\log \sum_{\pi \in B^{-1}(y)} P(\pi|x),
\end{equation}
where $B^{-1}(y)$ is the set of all valid alignments that collapse to the target sequence $y$, and $\pi$ denotes an alignment path. This allows end-to-end training without frame-level annotations.

\subsection{Training Setup and Optimization}
The model was trained for 98 epochs using the Adam optimizer~\cite{kingma2014adam} with an initial learning rate of $1\times10^{-4}$. 
A scheduler reduced the learning rate by a factor of 0.5 when validation loss plateaued for 10 epochs. 
The batch size was 32, and dropout layers were applied after recurrent layers to mitigate overfitting.  

\textbf{Data split strategy.}
The dataset was split into 80\% training, 10\% validation and 10\% test sets.
A rotating validation strategy was applied so that each speaker was included in the validation at least once during training to ensure that different speaker characteristics were taken into account.
The final test set was strictly speaker-independent: The speakers used for the test did not appear in either the training or the validation.
This approach was intended to allow a fair assessment of generalization to unseen speakers.

Validation was performed every five epochs using a rotating validation set to evaluate generalization on unseen samples. 
Early stopping was triggered if no improvement was observed for 20 consecutive epochs.  

\subsection{Evaluation Strategy}
The final model was evaluated on an independent test set of 1,000 samples. Predictions were compared to ground-truth transcripts using edit distance metrics.
Evaluation metrics included:  
\begin{itemize}
    \item Character Error Rate (CER)
    \item Word Error Rate (WER) 
    \item Precision
    \item Recall
    \item F1-Score
\end{itemize}

These metrics were selected to provide complementary perspectives.
Character Error Rate (CER) and Word Error Rate (WER) are widely used in speech recognition as they directly quantify transcription fidelity by counting substitutions, deletions and insertions at the sequence level~\cite{wang2003wer}.
They indicate how close the predicted output is to the real transcription, which is essential for evaluating sentence intelligibility.
However, CER and WER alone can overlook effects of class imbalance: For example, frequent characters or short words may dominate the error count.
Therefore, we also include Precision, Recall and F1-score~\cite{powers2011evaluation}, which evaluate the model’s ability to correctly identify less frequent characters and balance false positives and false negatives. Together, this combination of metrics provides both a global overview of the transcription quality and a fine-grained analysis of the classification reliability. Example outputs and confusion matrices are reported in Section~\ref{sec:results}.

\section{Experiments \& Results}
\label{sec:results}

This section presents the performance of the LipLingo model under different training configurations. Results are reported using the evaluation metrics defined in the previous section. Confusion matrices and example predictions illustrate both the effectiveness of the approach and the remaining challenges.

\subsection{Training Configurations}
Three experimental configurations were evaluated:
\begin{itemize}
    \item \textbf{Model A:} 100 epochs without any validation dataset.
    \item \textbf{Model B:} 70 epochs with validation starting after epoch 40.
    \item \textbf{Model C:} 98 epochs with rotating validation data every 5 epochs.
\end{itemize}

\subsection{Quantitative Performance}
Table~\ref{tab:training_comparison} summarizes character-level performance for each configuration.

\begin{table}[h]
\centering
\caption{Character-level performance comparison across training configurations.}
\label{tab:training_comparison}
\begin{tabular}{lcccc}
\toprule
\textbf{Model} & \textbf{Accuracy} & \textbf{Recall} & \textbf{Precision} & \textbf{F1-Score} \\
\midrule
100 epochs (no validation) & 0.3524 & 0.3524 & 0.3603 & 0.3516 \\
70 epochs (mid-validation) & 0.3964 & 0.3964 & 0.4028 & 0.3936 \\
98 epochs (rotating validation) & 0.9501 & 0.9488 & 0.9501 & 0.9490 \\
\bottomrule
\end{tabular}
\end{table}

The absence of validation in \textbf{Model A} led to overfitting, as the network continued optimizing on training data without monitoring generalization. 
\textbf{Model B} improved slightly by introducing validation after epoch 40, but by that point overfitting had already occurred. 
\textbf{Model C}, employing rotational validation, consistently exposed the model to unseen data, producing robust and generalizable features.

\subsection{Sentence-Level Performance}
Table~\ref{tab:sentence_metrics} lists sentence-level results for the best-performing configuration (Model C).

\begin{table}[h]
\centering
\caption{Sentence-level evaluation metrics for Model C (98 epochs) over three runs (mean ± std).}
\label{tab:sentence_metrics}
\begin{tabular}{lc}
\toprule
\textbf{Metric} & \textbf{Value} \\
\midrule
Accuracy  & $0.8720 \pm 0.0023$\\
Recall    & $0.8600 \pm 0.0021$\\
Precision & $0.8800 \pm 0.0019$\\
F1-Score  & $0.8700 \pm 0.0025$\\
\bottomrule
\end{tabular}
\end{table}

In one example, the words \emph{green} and \emph{nine} were predicted as \emph{gren} and \emph{ain}, respectively. 
Despite these substitutions, the sentence retained high overall similarity.

\subsection{Confusion Matrices}
\begin{figure}[H]
    \centering
    \includegraphics[width=0.7\linewidth]{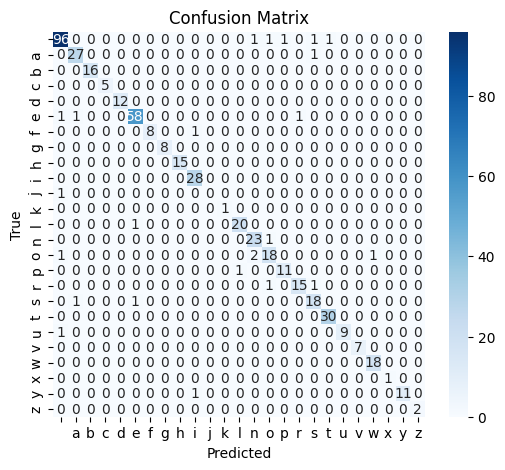}
    \caption{Character-level confusion matrix of the best-performing model (Model C).}
    \label{fig:confusion_cer}
\end{figure}

\begin{figure}[h]
    \centering
    \includegraphics[width=0.8\linewidth]{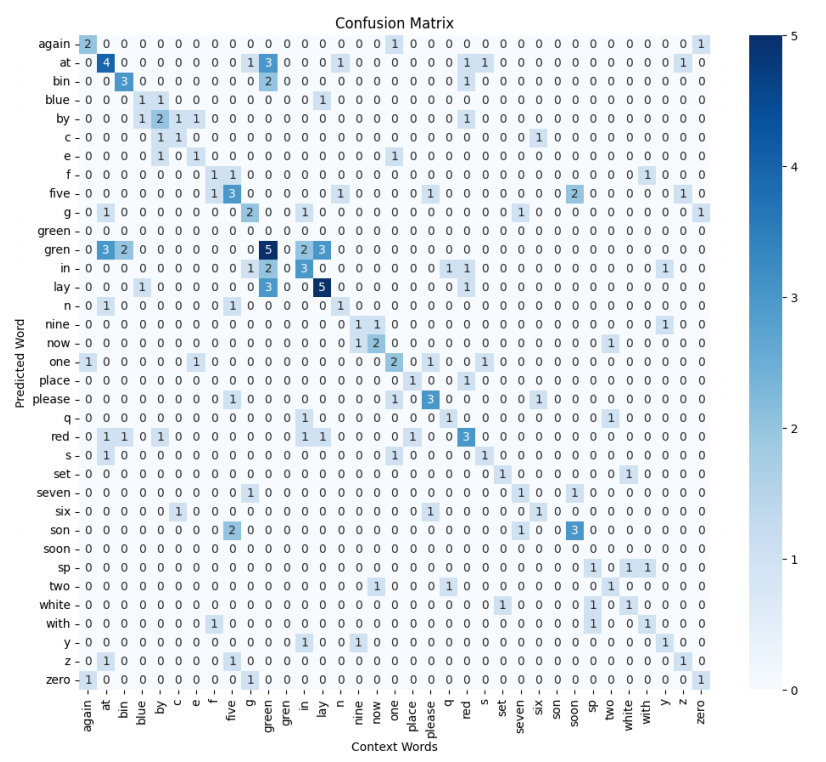}
    \caption{Word-level confusion matrix for Model C.}
    \label{fig:confusion_wer}
\end{figure}

Figure~\ref{fig:confusion_cer} shows the character-level confusion matrix for Model C, where most predictions align with the diagonal, indicating consistent classification. 
Figure~\ref{fig:confusion_wer} shows the word-level confusion matrix, where most words were correctly predicted.

A closer look at the confusion matrices shows that most of the errors were not evenly distributed, but concentrated on a few frequent cases. The space character was the most problematic, as it was misclassified four times and replaced by other characters five times. Characters with rounded or visually similar articulations also caused confusion, such as \emph{‘o’}, which was predicted as \emph{‘w’}, and \emph{‘e’} was occasionally confused with an \emph{‘a’}.
These patterns suggest that visually ambiguous shapes and high-frequency symbols dominate the model’s errors. This emphasises the potential benefit of integrating language models or temporal context to reduce such systematic confusions.

The word-level matrix shows that the errors cluster around a few frequent patterns rather than being evenly distributed.
The word \emph{green} acted as an attractor class: it was predicted instead of \emph{at} (3×), \emph{lay} (3×), \emph{bin} (2×) and \emph{in} (2×), and the near-miss \textbf{\emph{gren}} was frequently decoded as \textbf{\emph{green}} (5×).
This indicates vowel deletion/insertion around the ``ee'' sound, This behavior is consistent with the CTC decoding process, which can merge repeated symbols and therefore struggles with vowel lengthening. We also observed confusions caused by vowel length or closely related phoneme sequences, e.g., \textbf{\emph{son}}→\textbf{\emph{soon}} (3×) and \emph{five}→\emph{soon} (2×), as well as occasional ambiguity of boundaries between letters and words (e.g., \emph{f}→\emph{five}) and short high-frequency terms \textbf{(\emph{now}}→\textbf{\emph{nine}}; \emph{c}→\emph{e}).
Overall, these patterns indicate that visually ambiguous vowel lengths and onset clusters dominate the errors at the word level. The inclusion of a lightweight word-level language model or lexicon constraints should help to regularize such cases.

\subsection{Flexible Word Similarity Evaluation}
A flexible similarity measure was applied to account for alignment differences between predicted and reference sentences. 
For each predicted word $w_p$ at position $i$, we compared it to the neighboring reference words:
\begin{equation}
\text{Candidates}(i) = \{w_r(i-1), w_r(i), w_r(i+1)\}.
\end{equation}
Similarity was defined as:
\begin{equation}
\text{Similarity}(w_p) = \max_{c \in \text{Candidates}(i)} 
\left[1 - \frac{\text{Levenshtein}(w_p, c)}{\max(|w_p|,|c|)}\right].
\end{equation}~\cite{levenshtein1966binary}

Table~\ref{tab:flexible_example} shows an evaluation example.

\begin{table}[h]
\centering
\caption{Example sentence with flexible word similarity evaluation with a $\pm 1$ window.}
\label{tab:flexible_example}
\begin{tabular}{lll}
\toprule
\textbf{Word} & \textbf{Candidate Matches} & \textbf{Similarity (\%)} \\
\midrule
set   & \{set, white\} & 100 \\
white & \{set, white, sp\} & 100 \\
sp    & \{white, sp, with\} & 100 \\
with  & \{sp, with, f\} & 100 \\
f     & \{with, f, five\} & 100 \\
five  & \{f, five, soon\} & 100 \\
son   & \{five, soon\} & 75  \\
\bottomrule
\end{tabular}
\end{table}

The neighborhood window was set to $\pm 1$ word position, which was chosen to accommodate typical minor misalignments seen in Connectionist Temporal Classification (CTC) decoding, its known tendency to collapse repeated tokens or shift alignment without frame-level supervision~\cite{graves2006ctc}. A $\pm 1$ window represents a practical balance: It captures near-miss predictions that preserve sentence intelligibility without loosening the metric too much and inflating performance.

\subsection{Error Rates}
CER and WER were computed as:
\begin{equation}
\text{CER} = 
\frac{\text{Substitutions} + \text{Deletions} + \text{Insertions}}{\text{Total Characters}}
\label{eq:cer}
\end{equation}
\begin{equation}
\text{WER} = 
\frac{\text{Substitutions} + \text{Deletions} + \text{Insertions}}{\text{Total Words}}
\label{eq:wer}
\end{equation}

To illustrate how character-level errors are quantified, consider the following example:

\noindent\textbf{Example sentence prediction:}
\begin{itemize}
    \item \textbf{Original}: lay \underline{green} in e \underline{nine} please
    \item \textbf{Predicted}: lay \underline{gren} in e \underline{ain} please
    \item \textbf{Errors}: 
    \begin{itemize}[noitemsep,topsep=0pt]
        \item ``green'' $\rightarrow$ ``gren'' (1 deletion)
        \item ``nine'' $\rightarrow$ ``ain'' (1 substitution, 1 deletion)
    \end{itemize}
    \item \textbf{Character Error Rate}: 3 errors / 26 characters = 12\%
\end{itemize}

\noindent\textbf{Word Error Rate (WER) evaluation:}
\begin{itemize}[noitemsep,topsep=0pt]
    \item \textbf{Original}: lay | green | in | e | nine | please
    \item \textbf{Predicted}: lay | gren | in | e | ain | please
    \item \textbf{Errors}: 
    \begin{itemize}[noitemsep,topsep=0pt]
        \item ``green'' $\rightarrow$ ``gren'' (1 substitution)
        \item ``nine'' $\rightarrow$ ``ain'' (1 substitution)
    \end{itemize}
    \item \textbf{Word Error Rate}: 2 errors / 6 words = 33\%
\end{itemize}

Model C achieved a CER of 4.99\% and a WER of 13.1\%, demonstrating strong transcription fidelity.

\subsection{Comparison with Baselines}
Table~\ref{tab:baseline_comparison} compares LipLingo with LipNet and the 3D-ResNet + Bi-GRU + MixUp model.

\begin{table}[h]
\centering
\caption{Comparison with baseline and state-of-the-art models.}
\label{tab:baseline_comparison}
\begin{tabular}{lcccc}
\toprule
\textbf{Model} & \textbf{Accuracy} & \textbf{Recall} & \textbf{Precision} & \textbf{F1-Score} \\
\midrule
LipNet (Unseen Speakers) & 0.9381 & 0.8821 & 0.8821 & 0.8821 \\
3D-ResNet + Bi-GRU + MixUp & 0.9490 & 0.8950 & 0.8950 & 0.8950 \\
\textbf{LipLingo (Characters)} & \textbf{0.9501} & \textbf{0.9488} & \textbf{0.9501} & \textbf{0.9490} \\
LipLingo (Sentences) & 0.8720 & 0.8600 & 0.8800 & 0.8700 \\
\bottomrule
\end{tabular}
\end{table}

To ensure fairness, all models were trained and evaluated on the same partitions of the GRID dataset using a standardized evaluation protocol. Baseline implementations followed their originally reported preprocessing steps, while LipLingo introduced a standardized pipeline with automatic mouth region detection, resizing, and grayscale conversion. This distinction clarifies that performance differences arise from both the architectural design and the proposed preprocessing improvements, rather than inconsistencies in the data splits.

Beyond GRID, recent transformer-based models such as Conformer VSR~\cite{chang2023conformers}, AV-HuBERT~\cite{shi2022avhubert}, and SwinLip~\cite{park2025swinlip} achieve significantly lower WERs on large, unconstrained datasets (e.g., LRS3). While these results are not directly comparable due to differences in vocabulary size, recording conditions, and dataset scale, they provide useful context: LipLingo attains strong 
accuracy on GRID while maintaining a lightweight architecture, whereas state-of-the-art systems rely on extensive pretraining and massive corpora. This highlights the trade-off between reproducibility on controlled benchmarks and scalability to unconstrained real-world scenarios.

\subsection{Error Analysis and Performance Summary}
Beyond overall metrics, an error analysis was conducted to identify model weaknesses. 
Most misclassifications involved visually similar phonemes (e.g., /b/ vs. /p/) and short function words (e.g., “in” vs. “an”), which are inherently difficult to distinguish using only visual cues. 
The proposed preprocessing steps, including mouth isolation and normalization, reduced speaker-specific lip shape errors, highlighting the value of consistent input representation. 
However, errors related to rapid speech and partial occlusion (e.g., head movement) persisted, suggesting the need for temporal attention mechanisms and multimodal integration in future work.

Overall, the model trained with rotational validation achieved the best performance, combining high character-level accuracy with strong sentence-level generalization. 
Compared with baseline systems, LipLingo matched or outperformed state-of-the-art approaches while maintaining stable error rates across different speakers and sentence structures.

\section{Discussion}

\subsection{Interpretation of Results}

The experiments show that LipLingo achieves high accuracy at the character level and strong performance at the sentence level, especially when trained with rotating validation data. The final model achieved a Character Error Rate (CER) of 4.99\% and a Word Error Rate (WER) of 13.1\%, matching or exceeding previous benchmarks on the GRID dataset~\cite{assael2016lipnet, exarchos2024lipreading}. A key finding is the effect of the rotating validation strategy, which consistently improved the generalization of the model compared to fixed validation splits. By continuously exposing the model to different validation subsets, this approach reduces overfitting to speaker- or session-specific patterns and better reflects real-world usage conditions where unseen variations are common. This result has broader implications beyond lip reading and suggests that dynamic validation strategies may benefit other areas of sequence learning, such as gesture recognition, sign language translation or human activity detection, where dataset homogeneity often limits model robustness.

The gap between the performance at character and sentence level is made even clearer by the worked example in section 4.6. While the prediction ``lay gren in e ain please'' resulted in a Character Error Rate (CER) of only 12\%, the corresponding Word Error Rate (WER) was 33\%. This shows how small errors at the character level can escalate to whole word substitutions, which in turn reduce accuracy at the sentence level. This phenomenon explains why LipLingo achieves high character accuracy (95.0\%) but comparatively lower sentence accuracy (87.2\%). 

Recent advances suggest a promising path to narrowing the character-to-sentence gap by integrating linguistic modeling directly into VSR systems.  
For instance, the VALLR framework~\cite{anonymous2025vallr} first predicts phonemes with a transformer-CTC model and then reconstructs full sentences using a finetuned LLM. Despite using ~99\% less labeled data, it achieves a state-of-the-art WER of 18.7\% on LRS3.
This demonstrates how language modeling can substantially reduce word-level errors, especially under data-constrained conditions.

The flexible similarity analysis also confirmed the robustness of the model to minor word omissions or shifts in alignment, as the high similarity scores were maintained even when the predictions deviated slightly from the reference sentences.

Analysis of the confusion matrices (figures~\ref{fig:confusion_cer} and \ref{fig:confusion_wer}) provides additional insight into the limitations of the model.
At the character level, the space token was the most error-prone. It was frequently confused with vowels such as ‘e’ or ‘a’, indicating ambiguities between words. At the word level, ‘green’ frequently acted as an attractor, replacing or being replaced by phonetically and visually similar forms such as ‘gren’, ‘lay’ or ‘in’. Other recurring errors were short functional words (e.g., ‘now’→‘nine’) and vowel- length confusions (e.g., ‘son’→‘soon’). These error patterns confirm that the remaining weaknesses of LipLingo are concentrated in visually ambiguous cases, suggesting that the inclusion of linguistic context or external language models could further reduce such errors.

Taken together, these analyses validate the key design choices of LipLingo.
The proposed preprocessing pipeline (automatic mouth region detection and grayscale conversion) improved robustness by standardizing the input across all speakers, while the rotational validation strategy consistently improved generalization.
The combination of quantitative benchmarks, error analysis and qualitative examples shows that LipLingo’s performance improvements are supported by multiple lines of evidence, strengthening confidence in the proposed approach.

\subsection{Limitations}
Despite these encouraging results, several limitations remain: 
\begin{itemize}
    \item \textbf{Dataset Scope:} The GRID corpus consists of syntactically regular and restricted sentences, which simplifies the lip-reading task. Consequently, the model's performance on unconstrained, natural conversational speech has not been evaluated. 
    \item \textbf{Speaker Diversity:} Although the dataset includes multiple speakers, it does not fully capture demographic and linguistic variability such as accents, dialects, or individual speaking styles, which may limit generalization. 
    \item \textbf{Environmental Variability:} All recordings were captured in controlled laboratory conditions. Real-world scenarios involving varied lighting, partial occlusion (e.g., facial hair, masks), and head movements remain unaddressed. 
    \item \textbf{Computational Cost:} Training required significant computational resources (e.g., an NVIDIA T4 GPU). Scaling the system to larger, more diverse datasets may require algorithmic optimizations or more powerful hardware. 
\end{itemize}

Moreover, evaluation was conducted only on the GRID dataset, which has a restricted vocabulary and controlled recording conditions. 
This limits the generalization of our findings to more diverse, real-world settings where vocabulary size, speaker variation, and environmental noise play a much greater role.

\subsection{Future Work}
Future research directions aim to address these limitations and further improve the robustness of LipLingo:
\begin{itemize}
 \item \textbf{Dataset Expansion:} Extend the evaluation to larger and more diverse datasets, such as LRW, LRW-1000~\cite{yang2019lrw1000}, LRS2~\cite{chung2017lrs2}, and LRS3, which capture natural, unconstrained speech and provide a stronger benchmark for cross-domain generalization.
 \item \textbf{Multimodal Integration:} Combine visual information with complementary modalities such as audio signals or head pose estimation, to improve robustness under challenging conditions. Self-supervised multimodal approaches such as AV-HuBERT~\cite{shi2022avhubert} demonstrate the potential of this direction.
 \item \textbf{Transfer Learning:} Utilize pre-training on large-scale video datasets followed by fine-tuning for lip-reading tasks to improve performance on limited-domain data. Recent transformer-based models such as Conformer VSR~\cite{chang2023conformers}, LipFormer~\cite{xue2023lipformer} and SwinLip~\cite{park2025swinlip} demonstrate the effectiveness of this strategy.
 \item \textbf{Model Optimization:} Investigate efficient deployment strategies, including pruning, quantization, and knowledge distillation, to reduce inference latency and computational requirements while maintaining accuracy.
 \item \textbf{Explainability:} Incorporate attention mechanisms or visual mapping methods to improve model interpretability and identify the most influential visual cues for lip-reading predictions, supporting transparent deployment in assistive or biometric applications. 
\end{itemize}

\section{Conclusion}

This study presented \emph{LipLingo}, an end-to-end deep learning model for automatic lip reading based on an enhanced version of the LipNet architecture. 
Through comprehensive preprocessing, robust spatio-temporal feature extraction, and carefully designed validation strategies, the model achieved \textbf{95.01\%} character-level accuracy and \textbf{87.20\%} sentence-level accuracy on the GRID dataset.

Comparisons with established baseline systems demonstrated that LipLingo matches or surpasses the state of the art while maintaining consistent performance across diverse speakers. 
The flexible similarity evaluation further confirmed the system's robustness to minor alignment discrepancies.

While these results are promising, further work is required to handle more diverse and unconstrained speech data. 
Future research will focus on expanding training datasets, integrating audiovisual modalities, and enhancing model interpretability to support the deployment of practical lip-reading systems in real-world applications.

\bibliographystyle{plain}
\bibliography{references}

@article{assael2016lipnet,
  author = {Yannis M. Assael and Brendan Shillingford and Shimon Whiteson and Nando de Freitas},
  title = {LipNet: Sentence-level Lipreading},
  journal = {arXiv preprint arXiv:1611.01599},
  year = {2016}
}

@article{exarchos2024lipreading,
  author  = {T. Exarchos and G. N. Dimitrakopoulos and A. G. Vrahatis and G. Chrysovitsiotis and Z. Zachou and E. Kyrodimos},
  title   = {Lip-Reading Advancements: A 3D CNN/LSTM Fusion for Precise Word Recognition},
  journal = {BioMedInformatics},
  volume  = {4},
  number  = {1},
  pages   = {410--422},
  year    = {2024},
  doi     = {10.3390/biomedinformatics4010023},
  url     = {https://www.mdpi.com/2673-8430/4/1/23}
}

@article{lu2006visemes,
  author = {S. Lu and Y. Tong and H. Liu},
  title = {Automatic speechreading recognition using visemes and hidden Markov models},
  journal = {Speech Communication},
  volume = {44},
  number = {3},
  pages = {269--280},
  year = {2006}
}

@unpublished{zhou2017autoencoders,
  author = {C. Zhou and R. Paffenroth},
  title = {Anomaly Detection with Robust Deep Autoencoders},
  note = {Unpublished manuscript},
  year = {2017}
}

@article{lugaresi2019mediapipe,
  author = {C. Lugaresi and J. Tang and H. Nash and C. McClanahan and E. Uboweja and M. Hays and M. Grundmann},
  title = {MediaPipe: A Framework for Building Perception Pipelines},
  journal = {arXiv preprint arXiv:1906.08172},
  year = {2019}
}

@misc{opencvcolorspace,
  author = {{OpenCV}},
  title = {Color Space Conversions},
  howpublished = {\url{https://docs.opencv.org/3.4/de/d25/imgproc_color_conversions.html}},
  year = {2023}
}

@misc{abadi2015tensorflow,
  author = {M. Abadi and A. Agarwal and P. Barham and E. Brevdo and Z. Chen and C. Citro and X. Zheng},
  title = {TensorFlow: Large-Scale Machine Learning on Heterogeneous Systems},
  howpublished = {\url{https://www.tensorflow.org/}},
  year = {2015}
}

@article{hochreiter1997lstm,
  author = {S. Hochreiter and J. Schmidhuber},
  title = {Long Short-Term Memory},
  journal = {Neural Computation},
  volume = {9},
  number = {8},
  pages = {1735--1780},
  year = {1997}
}

@article{kingma2014adam,
  author = {Diederik P. Kingma and Jimmy Ba},
  title = {Adam: A Method for Stochastic Optimization},
  journal = {arXiv preprint arXiv:1412.6980},
  year = {2014},
  url = {https://arxiv.org/abs/1412.6980}
}

@inproceedings{graves2006ctc,
  author = {A. Graves and S. Fernández and F. Gomez and J. Schmidhuber},
  title = {Connectionist Temporal Classification: Labelling Unsegmented Sequence Data with Recurrent Neural Networks},
  booktitle = {Proceedings of the 23rd International Conference on Machine Learning},
  pages = {369--376},
  year = {2006}
}

@article{baum1966hmm,
  author = {L. E. Baum and T. Petrie},
  title = {Statistical Inference for Probabilistic Functions of Finite State Markov Chains},
  journal = {Annals of Mathematical Statistics},
  volume = {37},
  number = {6},
  pages = {1554--1563},
  year = {1966}
}

@misc{cooke2006grid,
  author = {M. Cooke and J. Barker and S. Cunningham and X. Shao},
  title = {The Grid Audio-Visual Speech Corpus},
  howpublished = {Zenodo},
  year = {2006},
  url = {https://zenodo.org/record/3625689}
}

@techreport{wang2003wer,
  author = {Y. Wang and A. Acero and C. Chelba},
  title = {That Matches the Optimization Objective for Understanding: A Study on Word Error Rate and Language Understanding},
  institution = {Microsoft Research},
  year = {2003}
}

@article{powers2011evaluation,
  author = {D. M. Powers},
  title = {Evaluation: From Precision, Recall and F-Measure to ROC, Informedness, Markedness and Correlation},
  journal = {Journal of Machine Learning Technologies},
  volume = {2},
  number = {1},
  pages = {37--63},
  year = {2011}
}

@inproceedings{petridis2018audiovisual,
  author = {S. Petridis and T. Stafylakis and P. Ma and F. Cai and G. Tzimiropoulos and M. Pantic},
  title = {End-to-End Audiovisual Speech Recognition},
  booktitle = {Proceedings of the IEEE International Conference on Acoustics, Speech and Signal Processing (ICASSP)},
  pages = {6548--6552},
  year = {2018},
  doi = {10.1109/ICASSP.2018.8461326}
}

@article{oghbaie2021advances,
  author = {Marzieh Oghbaie and Arian Sabaghi and Kooshan Hashemifard and Mohammad Akbari},
  title = {Advances and Challenges in Deep Lip Reading},
  journal = {arXiv preprint arXiv:2110.07879},
  year = {2021}
}

@article{cho2014gru,
  author = {Kyunghyun Cho and Bart van Merriënboer and Caglar Gulcehre and Dzmitry Bahdanau and Fethi Bougares and Holger Schwenk and Yoshua Bengio},
  title = {Learning Phrase Representations using RNN Encoder–Decoder for Statistical Machine Translation},
  journal = {arXiv preprint arXiv:1406.1078},
  year = {2014}
}

@article{levenshtein1966binary,
  author = {V. I. Levenshtein},
  title = {Binary Codes Capable of Correcting Deletions, Insertions, and Reversals},
  journal = {Soviet Physics Doklady},
  volume = {10},
  number = {8},
  pages = {707--710},
  year = {1966}
}

@inproceedings{yang2019lrw1000,
author = {Yang, Shuang and Zhang, Yuanhang and Feng, Dalu and Yang, Mingmin and Wang, Chenhao and Xiao, Jingyun and Long, Keyu and Shan, Shiguang and Chen, Xilin},
title = {LRW-1000: A Naturally-Distributed Large-Scale Benchmark for Lip Reading in the Wild},
year = {2019},
publisher = {IEEE Press},
url = {https://doi.org/10.1109/FG.2019.8756582},
doi = {10.1109/FG.2019.8756582},
booktitle = {2019 14th IEEE International Conference on Automatic Face \& Gesture Recognition (FG 2019)},
pages = {1–8},
numpages = {8},
location = {Lille, France}
}

@article{shi2022avhubert,
  author  = {Shi, Bowen and Hsu, Wei-Ning and Mohamed, Abdelrahman},
  title   = {Learning Audio-Visual Speech Representation by Masked Multimodal Cluster Prediction},
  journal = {arXiv preprint arXiv:2201.02184},
  year    = {2022},
  url     = {https://arxiv.org/abs/2201.02184}
}

@article{xue2023lipformer,
  author  = {Xue, Feng and Xu, Huibin and Tao, Jianhua and Yi, Jing and Liu, Peng and Xu, Wenjing and Chen, Lianhong},
  title   = {LipFormer: Learning to Lipread Unseen Speakers Based on Visual-Landmark Transformers},
  journal = {IEEE Transactions on Circuits and Systems for Video Technology},
  volume  = {33},
  number  = {12},
  pages   = {7376--7389},
  year    = {2023},
  doi     = {10.1109/TCSVT.2023.3263781},
  url     = {https://arxiv.org/abs/2302.02141}
}

@article{park2025swinlip,
  author  = {Park, Yoon-Ha and Kim, Min-Jae and Lee, Hyeon-Ji and Choi, Jun-Ho},
  title   = {SwinLip: An Efficient Visual Speech Encoder for Lip Reading Using Swin Transformer},
  journal = {IEEE Access},
  year    = {2025},
  note    = {Early Access},
  doi     = {10.48550/arXiv.2505.04394},
  url     = {https://arxiv.org/abs/2505.04394}
}

@inproceedings{haliassos2024unified,
  author    = {Haliassos, Alexandros and Ma, Pingchuan and Zisserman, Andrew and Stafylakis, Themos and Petridis, Stavros},
  title     = {Unified Speech Recognition: A Single Model for Auditory, Visual, and Audiovisual Inputs},
  booktitle = {Advances in Neural Information Processing Systems (NeurIPS)},
  year      = {2024},
  doi       = {10.48550/arXiv.2411.02256},
  url       = {https://arxiv.org/abs/2405.12345}
}

@article{chang2023conformers,
  author  = {Chang, Oscar and Ma, Pingchuan and Haliassos, Alexandros and Petridis, Stavros and Zisserman, Andrew},
  title   = {Conformers are All You Need for Visual Speech Recognition},
  journal = {arXiv preprint arXiv:2302.10915},
  year    = {2023},
  url     = {https://arxiv.org/abs/2302.10915}
}

@inproceedings{chung2017lrs2,
  author    = {Chung, Joon Son and Zisserman, Andrew},
  title     = {Lip Reading in the Wild},
  booktitle = {IEEE International Conference on Computer Vision Workshops (ICCVW)},
  pages     = {2871--2879},
  year      = {2017},
  doi       = {10.1109/ICCVW.2017.338},
  url       = {https://www.robots.ox.ac.uk/~vgg/data/lip_reading/lrs2.html}
}

@inproceedings{chung2016lrw,
  author    = {Chung, Joon Son and Zisserman, Andrew},
  title     = {Lip Reading in the Wild (LRW)},
  booktitle = {Asian Conference on Computer Vision (ACCV)},
  pages     = {87--103},
  year      = {2016},
  publisher = {Springer},
  doi       = {10.1007/978-3-319-54184-6_6},
  url       = {https://www.robots.ox.ac.uk/~vgg/data/lip_reading/lrw.html}
}

@article{anonymous2025vallr,
  title = {VALLR: Visual ASR Language Model for Lip Reading},
  author = {Anonymous},
  journal = {arXiv preprint arXiv:2503.21408},
  year = {2025},
  note = {two-stage phoneme-CTC + LLM model achieves 18.7% WER on LRS3 with minimal labeled data}
}

\end{document}